%% file: emnlp2020-templates/main.tex
\pdfoutput=1

\documentclass[11pt]{article}

\usepackage{acl}
\usepackage{times}
\usepackage{latexsym}
\usepackage[T1]{fontenc}

\usepackage[utf8]{inputenc}

\usepackage{microtype}

\usepackage{hyperref}
\usepackage{booktabs}
\usepackage{graphicx}
\graphicspath{{./imgs/}}
\usepackage{footmisc}
\usepackage[most]{tcolorbox}
\usepackage{enumitem}

\usepackage{microtype}
\usepackage{arydshln}
\usepackage{caption}
\usepackage{subcaption}
\usepackage{array, makecell} %
\usepackage{footmisc}
\usepackage{alltt}
\usepackage{floatrow}
\usepackage{pifont}
\newcommand{\cmark}{\ding{51}}%
\newcommand{\xmark}{\ding{55}}%

\usepackage{tabularx}
\usepackage{adjustbox}
\usepackage{multirow}
\usepackage{multirow, colortbl, caption}
\definecolor{lightblue}{RGB}{232, 244, 248}
\definecolor{lightpink}{RGB}{254, 238, 237}

\definecolor{bluelink}{RGB}{0,113,188}
\definecolor{greenlink}{RGB}{0,188,113}
\definecolor{darkblue}{rgb}{0,0,0.55} %

\usepackage{tikz}

\usepackage{collcell}

\usepackage{etoolbox}

\newtoggle{inTableHeader}
\toggletrue{inTableHeader}
\newcommand*{\StartTableHeader}{\global\toggletrue{inTableHeader}}%
\let\OldTabular\tabular%
\let\OldEndTabular\endtabular%
\renewenvironment{tabular}{\StartTableHeader\OldTabular}{\OldEndTabular\StartTableHeader}%

\newcommand*{\MinNumber}{-1.0}%
\newcommand*{\MidNumber}{0.0} %
\newcommand*{\MaxNumber}{1.0}%

\newcommand{\ApplyGradient}[1]{%
  \iftoggle{inTableHeader}{#1}{
    \ifdim #1 pt > \MidNumber pt
        \pgfmathsetmacro{\PercentColor}{max(min(100.0*(#1 - \MidNumber)/(\MaxNumber-\MidNumber),100.0),0.00)} %
        \hspace{-0.33em}\colorbox{yellow!\PercentColor!blue}{#1}
    \else
        \pgfmathsetmacro{\PercentColor}{max(min(100.0*(\MidNumber - #1)/(\MidNumber-\MinNumber),100.0),0.00)} %
        \hspace{-0.33em}\colorbox{blue!\PercentColor!blue}{#1}
    \fi
  }}
\newcolumntype{R}{>{\collectcell\ApplyGradient}c<{\endcollectcell}}

\input{emnlp2020-templates/math-com}

\usepackage[nameinlink]{cleveref}
\crefformat{section}{\S#2#1#3} 
\crefname{algorithm}{Alg.}{Algs.}
\crefname{table}{Table}{Tables}
\crefformat{subsection}{\S#2#1#3}
\Crefname{equation}{Eq.}{Eqs.}
\Crefname{figure}{Figure}{Figures}
\usepackage{float}

\usepackage[colorinlistoftodos,prependcaption,textsize=tiny]{todonotes}

\definecolor{darkgreen}{rgb}{0,0.5,0}  

\usepackage{soul}

\usepackage{float}

\usepackage{multirow}
\usepackage{hhline}

\definecolor{chartqapro1}{RGB}{30,160,220} 
\definecolor{chartqapro2}{RGB}{50,200,100} 

\title{
\textbf{VisEditBench: Can Vision-Language Models Edit Visualization Code from Multimodal Feedback?}
}

\author{
\textbf{Mizanur Rahman}\textsuperscript{\textdaggerdbl}
\thanks{Corresponding authors: \{mizanurr,enamulh\}@yorku.ca},
\textbf{Arshia Azimlu}\textsuperscript{\textdaggerdbl}
\thanks{These authors contributed equally.}, \\
\textbf{Shadikur Rahman}\textsuperscript{\textdaggerdbl}\footnotemark[2],
\textbf{Md Tahmid Rahman Laskar}\textsuperscript{\textdaggerdbl}\footnotemark[2],
\textbf{Amran Bhuiyan}\textsuperscript{\textdaggerdbl}\footnotemark[2], \\
\textbf{Shafiq Joty}\textsuperscript{\textdollar,\textparagraph},
\textbf{Enamul Hoque Prince}\textsuperscript{\textdaggerdbl}\footnotemark[1]
\\[2pt]
\textsuperscript{\textdaggerdbl}York University \\
\textsuperscript{\textdollar}Nanyang Technological University \quad
\textsuperscript{\textparagraph}Salesforce AI Research
}
\begin{document}
\maketitle


\begin{abstract} 
 \vspace{-1mm}

Vision-language models (VLMs) have shown strong capabilities in generating visualization code from textual or visual specifications. However, real-world visualization authoring is inherently iterative: users frequently revise existing visualizations to repair flawed charts or adapt them to desired styles. Existing benchmarks primarily evaluate generation from scratch, leaving visualization code editing from multimodal feedback largely unexplored. We introduce \textbf{VisEditBench}, a benchmark of 1,395 human-annotated visualization code-editing tasks grounded in realistic visualization workflows and failure cases. VisEditBench covers two practical settings: \textit{feedback-guided repair}, where models revise visualization code using buggy or marked charts together with textual feedback, and \textit{reference-guided restyling}, where models modify code to match a target chart image. Evaluating 20 state-of-the-art VLMs reveals that visualization code editing remains challenging: Claude-4.6-Sonnet achieves the best overall pass rate of 74.46\%, while most open-source models remain below 50\%. Performance is particularly weak on visually grounded style adaptation, where Claude-4.6-Sonnet achieves only 55.71\%. To establish a strong baseline, we further propose \textbf{VisEditAgent}, a render-grounded editing framework that iteratively generates, executes, validates, and refines candidate edits. Built on GPT-4o, VisEditAgent improves overall pass rate from 55.75\% to 67.99\%, demonstrating the importance of render-grounded feedback for faithful visualization editing. We will release VisEditBench at \url{https://github.com/vis-nlp/VisEditBench}.

\end{abstract}
\vspace{-3mm}

\section{Introduction}

\begin{figure}[t]
    \vspace{-3mm}
    \centering
    \includegraphics[width=\textwidth]{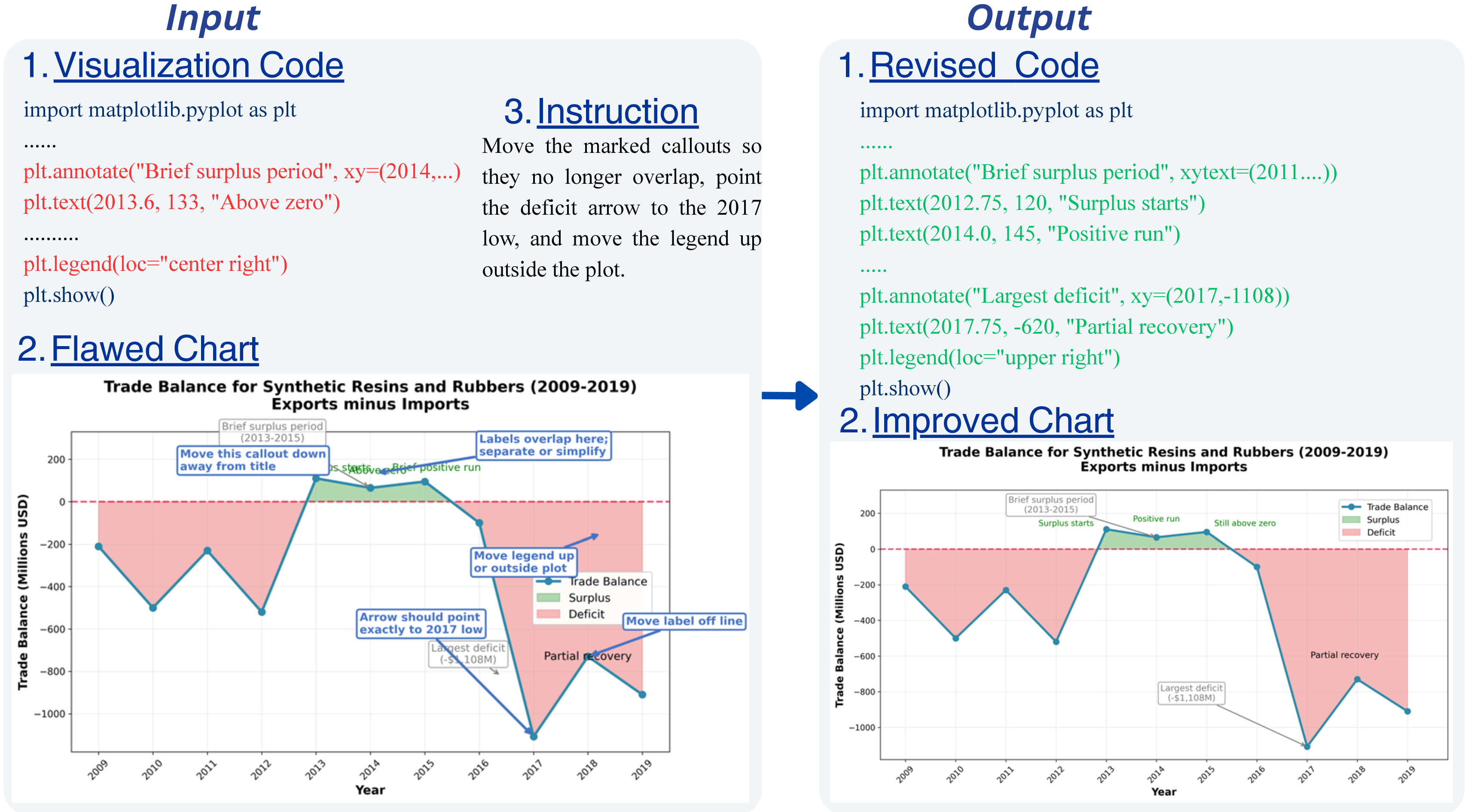}
    \caption{Overview of a VisEditBench task. Given visualization code, visual feedback or a reference chart, and a language instruction, the model must generate revised executable code that repairs or restyles the visualization while preserving the underlying data semantics.}
    \label{fig:visedit_example1}
    \vspace{-3mm}
\end{figure}
Data visualizations play a central role in modern data analysis and communication, enabling data scientists, business analysts, journalists, policymakers, and researchers to explore patterns, communicate insights, and support decision-making across domains \cite{hoque2024natural, rahman2025llm}. To support increasingly complex visualization workflows, recent advances in vision-language models (VLMs) have driven rapid progress in automated visualization, including chart generation from natural-language instructions \cite{rahman2025text2vis, maddigan2023chat2vis} and visualization code reconstruction from chart images \cite{yang2025chartmimic, wu2025plot2code}.

Yet real-world visualization authoring rarely ends with the first generated chart. Even when models produce executable visualizations, the outputs often fail to fully match user intent, visual preferences, accessibility requirements, or publication constraints \cite{rahman2025text2vis, chen2024viseval}. As a result, users frequently inspect rendered charts, revise the underlying code, and rerender the output through iterative refinement. In many workflows, charts must also be adapted to match preferred visual styles or reference designs, since default visualizations are often stylistically homogeneous and poorly aligned with communicative goals.  Figure~\ref{fig:visedit_example1} illustrates a typical repair workflow: although the user already has visualization code and a rendered chart, the output remains visually flawed; labels overlap, annotations obscure important regions, and visual emphasis is misplaced. Even seemingly minor edits can trigger cascading layout and styling changes, requiring repeated rounds of debugging and visual inspection.
This repair-and-restyling cycle is one of the most common and time-consuming bottlenecks in practical visualization workflows~\cite{harper2017converting}. Unlike one-shot generation, these workflows require localized edits that preserve data semantics, layout structure, encodings, and analytical meaning while modifying only the desired visual properties. 
Visualization authoring is therefore fundamentally an \textit{iterative multimodal editing problem}. Solving this problem requires grounded multimodal reasoning. A model must determine \textit{what} is visually wrong, \textit{which} code regions should change, \textit{how} the rendered chart should evolve, and \textit{which} aspects of the original visualization must remain preserved. This demands substantially richer reasoning than standard code generation or chart reconstruction.

Despite rapid progress in VLMs, existing benchmarks largely overlook this setting. Text-to-visualization benchmarks mainly evaluate chart generation from natural-language queries or data tables \cite{luo2021nvbench, liu2021advisor, rahman2025text2vis, chen2024viseval}, while chart-to-code benchmarks evaluate code reconstruction from chart images \cite{yang2025chartmimic, wu2025plot2code}. Neither captures a common workflow where users already have visualization code and want models to iteratively \textit{edit} rendered visualizations using multimodal feedback or reference charts. More general multimodal software-engineering benchmarks \cite{yang2024swe} study visual bug fixing, but do not address visualization-specific challenges such as repairing misleading encodings, resolving cluttered layouts, or matching reference chart styles while preserving data semantics.

To address this gap, we introduce \textbf{VisEditBench}, a benchmark for evaluating visualization code editing from multimodal feedback. Each task provides existing visualization code, a rendered chart, and a natural-language editing instruction, and requires the model to generate revised executable code that produces the desired visualization. VisEditBench supports two practical editing settings. In \textit{feedback-guided repair}, models receive buggy or human-marked charts together with textual feedback and must repair the visualization while preserving its intended meaning (Fig. \ref{fig:visedit_example1}). In \textit{reference-guided restyling}, models receive a target chart image and must adapt the original visualization to match the reference style without altering the underlying data semantics. Across these settings, VisEditBench contains 1,395 human-annotated tasks grounded in realistic visualization issues collected from Stack Overflow, Matplotlib and Vega-Lite issue reports, and diverse real-world visualization datasets.

Beyond executability, successful visualization editing requires edits that are visually faithful, semantically precise, and 
effective. To support fine-grained analysis, VisEditBench introduces a taxonomy of eight visualization-editing intents spanning correctness repair, readability improvement, style adaptation, consistency harmonization, robustness improvement, structural transformation, constraint satisfaction, and style-aware repair (Fig.~\ref{fig:visedit_examples}). Together, these settings capture realistic visualization workflows requiring visual reasoning, code understanding, 
and multimodal grounding.

We evaluate 20 state-of-the-art VLMs on VisEditBench in a zero-shot setting using executability, task accuracy, readability and clarity, visual quality, visual similarity, and final pass rate. Results show that visualization editing remains challenging: Claude-4.6-Sonnet achieves the best overall pass rate at 74.46\%, while most open-source models remain below 50\%. Performance is particularly weak on visually grounded style adaptation, where even Claude-4.6-Sonnet achieves only 55.71\%. To further study how rendering feedback can improve editing, we introduce \textbf{VisEditAgent}, a render-grounded baseline, 
that generates multiple candidate revisions, executes and renders them, visually validates the outputs, and iteratively refines the selected solution. Using GPT-4o as the base model, VisEditAgent improves overall pass rate from 55.75\% to 67.99\%, demonstrating that render-grounded feedback is critical for visually faithful visualization editing.

In summary, our contributions include:
\Ni \textbf{VisEditBench}, a benchmark of 1,395 human-annotated visualization code-editing tasks covering feedback-guided repair and reference-guided restyling across diverse visualization-editing intents; \Nii a structured \textbf{evaluation framework} measuring executability, task accuracy, chart readability and clarity, visual quality, visual similarity, and final pass rate;
\Niii \textbf{zero-shot evaluations} of 20 state-of-the-art VLMs, revealing major limitations in executable and visually faithful chart editing; and
\Niv \textbf{VisEditAgent}, a render-grounded baseline that improves visualization editing through execution feedback, visual validation, and iterative refinement.

\section{Related Work}
\textbf{Text-to-Visualization} Existing visualization benchmarks and systems have primarily studied how models generate~\cite{rahman2025text2vis}, understand~\cite{hoque2022chartquestionansweringstate}, or reconstruct charts~\cite{wu2025plot2code} from natural-language queries, tabular data, chart images, and analytical intents (Tab.~\ref{tab:benchmark_comparison}). 
WikiSQL~\cite{zhong2017seq2sql} supports natural-language-to-SQL parsing over tables, while nvBench~\cite{luo2021nvbench} extends this setting to large-scale cross-domain NL2VIS by synthesizing paired natural-language queries and visualization specifications from NL-to-SQL benchmarks. 
Later systems explore visualization recommendation, specification generation, and code generation from free-form analytical queries~\cite{liu2021advisor, dibia2019data2vis, narechania2020nl4dv, song2022rgvisnet, maddigan2023chat2vis}. 
Recent LLM-based benchmarks, including Text2Vis~\cite{rahman2025text2vis} and VisEval~\cite{chen2024viseval}, evaluate whether large language models can generate visualizations across diverse real-world datasets. ChartBench~\cite{xu2023chartbench} focuses on chart comprehension and visual reasoning, while ChartLlama~\cite{han2023chartllama}, ChartMimic~\cite{yang2025chartmimic}, and Plot2Code~\cite{wu2025plot2code} test chart reasoning or visualization-code reconstruction from chart images.  

However, existing benchmarks do not evaluate iterative visualization editing from multimodal feedback, where users revise existing visualization code using rendered feedback or reference charts while preserving data semantics (Tab. \ref{tab:benchmark_comparison}). VisEditBench addresses this gap by evaluating visualization code editing from multimodal feedback through feedback-guided repair and reference-guided restyling tasks.

\vspace{-2mm}
\paragraph{Multimodal Code Generation.}
Recent work has extended code generation to multimodal settings, where models generate or edit code from visual inputs \cite{li2024mmcode, si2025design2code, wu2025plot2code}. Early UI-to-code systems translated screenshots or sketches into executable interfaces \cite{beltramelli2018pix2code, robinson2019sketch2code}, while later work improved screenshot-to-code generation using vision-code architectures and visual feedback \cite{soselia2023learning}. More recent benchmarks study visually grounded programming tasks, including diagram-based programming (MMCode~\cite{li2024mmcode}), screenshot-to-webpage generation (Design2Code~\cite{si2025design2code}), SVG editing (SVGEditBench~\cite{nishina2024svgeditbench}), and visual software bug fixing (SWE-bench Multimodal~\cite{yang2024swe}). In contrast, VisEditBench focuses on editing and restyling existing visualization code from multimodal feedback while preserving data semantics.

\begin{figure*}[ht]
    \vspace{-10mm}
    \centering
    \includegraphics[width=\textwidth]{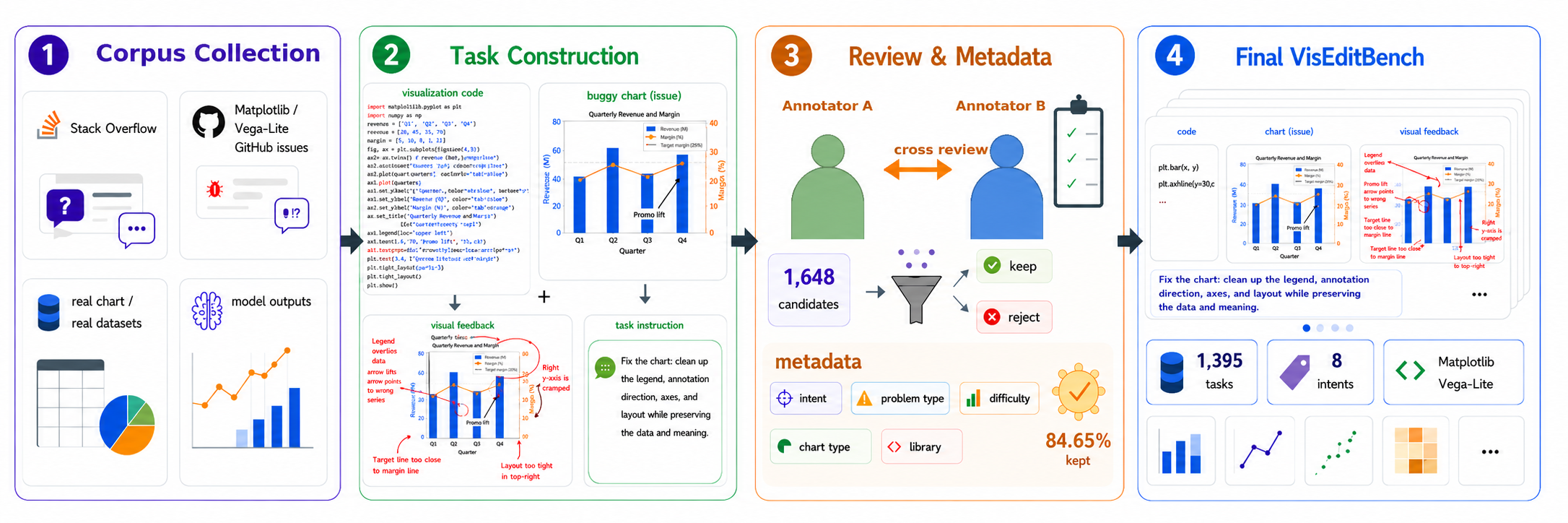}
    \caption{Overview of the VisEditBench construction pipeline. We collect real visualization issues and real-world chart/data examples, construct multimodal editing tasks, cross-review candidates for quality, and retain the final benchmark with metadata on editing intents, problem types, chart types, and visualization libraries.}
    \label{fig:visedit_flow}
    \vspace{-5mm}
\end{figure*} 

\section{\textsc{VisEditBench}}
\input{emnlp2020-templates/Dataset}
\section{Methodology}

\input{emnlp2020-templates/Models}

\vspace{-1mm}
\input{emnlp2020-templates/Evaluation}
\vspace{-1mm}
\section{Conclusion}
We introduce \textbf{VisEditBench}, the first benchmark for visualization code editing from multimodal feedback, moving beyond the one-shot chart-generation focus of prior work. Across 1,395 human-annotated tasks spanning feedback-guided repair and reference-guided restyling, our evaluation reveals a substantial gap between generating executable charts and performing visually faithful visualization editing, particularly for visually grounded and coordinated edits requiring layout, style, and semantic preservation. As a strong baseline, we include  
\textbf{VisEditAgent}, a render-grounded framework that iteratively generates, executes, validates, and refines candidate edits using visual feedback. This framework substantially improves editing performance, demonstrating that reliable visualization editing requires iterative multimodal reasoning rather than single-pass code generation. Together, VisEditBench and VisEditAgent establish a strong foundation for advancing visually grounded, feedback-aware, and user-aligned visualization authoring systems.

\section*{Limitations}
VisEditBench currently covers two visualization libraries, Matplotlib and Vega-Lite. These libraries represent both imperative and declarative visualization paradigms and support many common chart-editing scenarios. However, the benchmark does not yet directly evaluate other widely used visualization ecosystems such as Plotly, D3.js, and ggplot2. Extending VisEditBench to these libraries would allow future work to study whether the observed model behaviors generalize across a broader range of visualization programming environments.

Although VisEditBench includes diverse chart types, editing intents, problem types, and difficulty levels, it is not intended to exhaustively cover every possible visualization-editing need. The benchmark emphasizes common editing scenarios such as layout adjustment, label and annotation repair, encoding correction, style adaptation, and reference-guided restyling. More specialized domains may involve additional constraints, domain conventions, or visualization types that are not fully represented in the current dataset.

\section*{Ethical Considerations}

VisEditBench is designed to support transparent and reproducible research on multimodal visualization code editing. The benchmark is constructed from publicly available visualization resources, including user-reported visualization issues and real-world chart/data examples, and is used only for research and evaluation purposes. During dataset construction, annotators retained only examples with concrete visualization code, clearly grounded visual feedback, and implementable editing goals. We used LLM assistance only for language editing and polishing, while all technical content, dataset construction, experiments, and analyses were reviewed and verified by the authors.

To improve data quality and reduce ambiguity, each candidate task was cross-reviewed before inclusion in the final benchmark. We removed examples that were underspecified, duplicated, unrealistic, insufficiently grounded in the chart image, or not clearly solvable through visualization code editing. The benchmark does not aim to collect or evaluate sensitive personal information; its focus is on visualization code, rendered charts, editing instructions, and structured metadata.

We maintained fairness in model comparisons by applying the same prompting protocol, execution environment, evaluation criteria, and scoring rubrics across open-source and closed-source models. For automatic evaluation, we used a fixed rubric and avoided self-evaluation bias by using a different evaluator when assessing outputs from the default evaluator model. We also report limitations and failure modes to help prevent overclaiming model capabilities. Overall, VisEditBench is intended to encourage reliable, visually grounded, and user-aligned visualization authoring systems. Finally, we used AI-based writing assistants only to improve the presentation of the paper.

\section*{Acknowledgements}
This research was supported by the Natural Sciences and Engineering Research Council (NSERC),
Canada, Canada Foundation for Innovation, Compute Canada, and the CIRC grant on Inclusive and
Accessible Data Visualizations and Analytics.

\bibliography{text2vis}
\newpage
\input{emnlp2020-templates/Appendix}

\end{document}

%% file: emnlp2020-templates/math-com.tex
\usepackage{amsmath}
\usepackage{amsfonts,bm}
\usepackage{xspace}

\newcommand{\Ni}{({\em i})~}
\newcommand{\Nii}{({\em ii})~}
\newcommand{\Niii}{({\em iii})~}
\newcommand{\Niv}{({\em iv})~}

\definecolor{mypink3}{cmyk}{0, 0.7808, 0.4429, 0.1412}

\makeatletter   
\newcommand{\sveryshortarrow}[1][3pt]{\mathrel{%
    \vcenter{\hbox{\rule[-.5\fontdimen8\scriptfont3]
               {\scriptratio\dimexpr#1\relax}{\fontdimen8\scriptfont3}}}%
   \mkern-4mu\hbox{\let\f@size\sf@size\usefont{U}{lasy}{m}{n}\symbol{41}}}}
\makeatother

\def\eqref#1{equation~\ref{#1}}

\def\1{\bm{1}}

\def\m1{{\bm{1}}}

\DeclareMathAlphabet{\mathsfit}{\encodingdefault}{\sfdefault}{m}{sl}
\SetMathAlphabet{\mathsfit}{bold}{\encodingdefault}{\sfdefault}{bx}{n}



%% file: emnlp2020-templates/Dataset.tex
\vspace{-2mm}

We introduce VisEditBench, a benchmark of 1,395 human-annotated visualization code-editing tasks from multimodal feedback. 



\begin{figure*}[t!]
    \vspace{-5mm}
    \centering
    \includegraphics[width=\textwidth]{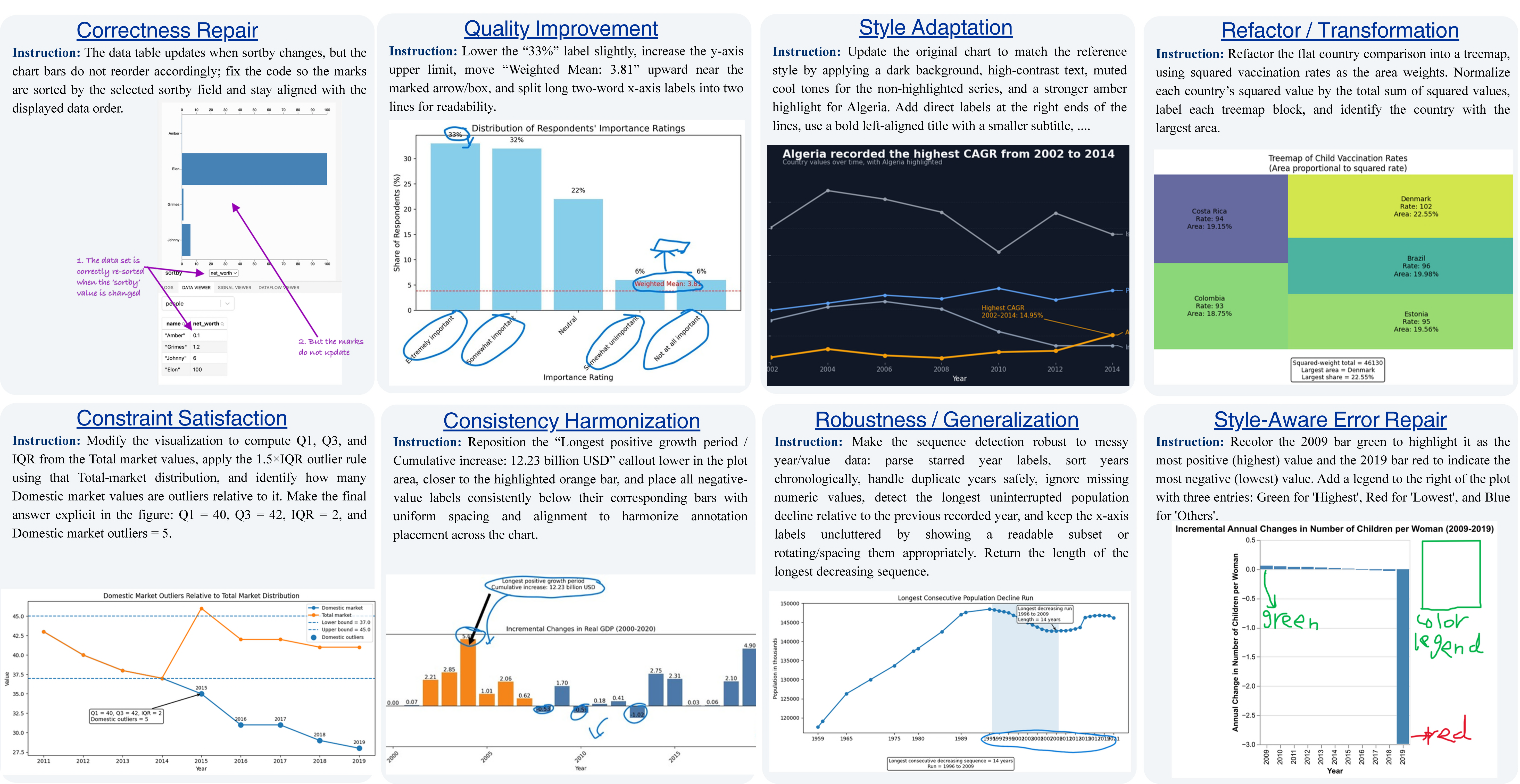}
    \caption{Examples from VisEditBench across eight editing intents. Each task pairs input visualization code with a buggy, human-marked, or reference chart image and a natural-language instruction; the model must output revised executable code that renders the desired visualization.}
    \label{fig:visedit_examples}
    \vspace{-3mm}
\end{figure*}
\vspace{-3mm}
\subsection{Data Collection}
We designed VisEditBench to capture realistic visualization editing workflows rather than synthetic chart-generation prompts. To ensure both realism and diversity, we combine two complementary sources: real user-reported visualization issues and model-generated chart failures on real-world datasets. Fig.~\ref{fig:visedit_flow} illustrates the  collection process.

First, we manually collected 120 real visualization-editing cases from Stack Overflow and Matplotlib/Vega-Lite GitHub issue discussions. We retained only examples containing a user-reported issue, visualization code, and a corresponding rendered buggy chart with clear visual problems requiring repair, improvement, or restyling. These naturally occurring cases include cluttered layouts, unreadable labels, malformed axes, misleading encodings, annotation failures, broken legends, and rendering inconsistencies. 


Second, we use the Text2Vis benchmark ~\cite{rahman2025text2vis}, whose tables and queries are drawn from realistic data sources such as Statista~\cite{statista}, Pew Research~\cite{pewresearch}, Our World in Data~\cite{owid}, and OECD~\cite{oecd}. We reran both frontier and open-source models on Text2Vis queries, executed the generated code, and manually inspected the rendered charts to identify visual, semantic, and presentation failures. We retained chart-code pairs only when the rendered output exhibited clear visualization errors or imperfect visual design, such as layout problems, incorrect encodings, unreadable labels, or poor presentation quality. These retained cases therefore reflect realistic examples where current models fail in zero-shot visualization generation. This process substantially increases diversity beyond naturally reported bugs, covering not only correctness repair, but also style adaptation, consistency harmonization, robustness improvement, structural transformation, and style-aware repair.



\subsection{Annotation and Quality Control}
From these sources, four annotators created an initial pool of 1,648 candidate editing tasks.  For each task, annotators wrote a natural-language editing instruction and assigned metadata including editing intent, problem type, visualization library, and chart type. Each candidate was then independently cross-reviewed by another annotator to verify that the task was visually grounded, realistic, clearly specified, and solvable through code edits. A candidate was retained only when the reviewer judged that the code, chart image, instruction, and metadata together formed a valid editing task with an observable visual target and a feasible code-level solution.

To measure annotation reliability, we computed inter-annotator agreement on the binary accept/reject decision across all candidates. Annotators agreed on 84.65\% of cases, indicating strong consensus on what constitutes a valid visualization editing task. We removed the remaining examples because they were ambiguous, insufficiently grounded in the image, duplicated, unrealistic, underspecified, or not clearly solvable through visualization code editing. This filtering process resulted in the final VisEditBench dataset of 1,395 high-quality tasks. Each retained instance contains existing visualization code, a chart image, textual feedback or editing instruction, visual feedback or a reference chart, and structured metadata for fine-grained analysis.

\begin{table*}[t!]
\vspace{-10mm}
\centering
\renewcommand{\arraystretch}{1.05}
\setlength{\tabcolsep}{4pt}
\setlength{\fboxsep}{5pt}

\caption{\small Statistics and diversity of VisEditBench. Counts and percentages are computed over the final 1,395 examples.}

\label{tab:viseditbench_statistics}

\begin{minipage}{0.985\textwidth}

\resizebox{\linewidth}{!}{%
\begin{tabular}{c c c c c c c c| c c c c c}
\multicolumn{8}{c}{\textbf{Editing Intent (Count / \%)}} &
\multicolumn{5}{c}{\textbf{Problem Type (Count / \%)}} \\
\cmidrule(lr){1-8} \cmidrule(lr){9-13}
\makecell{Correct.\\Repair} &
\makecell{Quality\\Improv.} &
\makecell{Robust.\\Gen.} &
\makecell{Style\\Adapt.} &
\makecell{Refactor/\\Transform.} &
\makecell{Consistency\\Harmon.} &
\makecell{Constraint\\Satis.} &
\makecell{Style-aware\\Repair} &
\makecell{Layout \&\\Geometry} &
\makecell{Annotation \&\\Labeling} &
\makecell{Encoding \&\\Mapping} &
\makecell{Data\\Transform.} &
Other \\
\midrule
\rowcolor[HTML]{EAF3FB}
529 / 37.9 & 446 / 32.0 & 116 / 8.3 & 70 / 5.0 &
63 / 4.5 & 59 / 4.2 & 52 / 3.7 & 49 / 3.5 &
802 / 57.5 & 348 / 24.9 & 156 / 11.2 & 63 / 4.5 & 26 / 1.9 \\
\bottomrule
\end{tabular}%
}

\vspace{2mm}

\resizebox{\linewidth}{!}{%
\begin{tabular}{c c| c c c |c c c c c| c c}
\multicolumn{2}{c}{\textbf{Cause Type (Count / \%)}} &
\multicolumn{3}{c}{\textbf{Complexity (Count / \%)}} &
\multicolumn{5}{c}{\textbf{Chart Type (Count / \%)}} &
\multicolumn{2}{c}{\textbf{Library (Count / \%)}} \\
\cmidrule(lr){1-2} \cmidrule(lr){3-5} \cmidrule(lr){6-10} \cmidrule(lr){11-12}
\makecell{Multi-cause} &
\makecell{Single-cause} &
Easy & Medium & Hard &
Bar & Line & Boxplot & Scatter & Other &
Matplotlib & Vega-Lite \\
\midrule
\rowcolor[HTML]{F3F8EE}
797 / 57.1 & 598 / 42.9 &
516 / 37.0 & 498 / 35.7 & 381 / 27.3 &
679 / 48.7 & 522 / 37.4 & 54 / 3.9 & 52 / 3.7 & 88 / 6.3 &
1156 / 82.9 & 239 / 17.1 \\
\bottomrule
\end{tabular}%
}

\end{minipage}%

\vspace{-3mm}
\end{table*}

\subsection{Dataset Diversity}
Table~\ref{tab:viseditbench_statistics} summarizes the diversity of VisEditBench. The dataset spans eight editing intents, with Correctness Repair (37.9\%) and Quality Improvement (32.0\%) forming the largest categories, reflecting the prevalence of misleading, cluttered, or visually ineffective charts in real workflows. VisEditBench also covers diverse problem types and difficulty levels. Layout \& Geometry issues are most common, followed by Annotation \& Labeling, Encoding \& Mapping, and Data Transformation. These categories include failures such as overlapping labels, malformed axes, incorrect encodings, broken annotations, and data-processing errors. Importantly, 57.1\% of tasks are multi-cause, requiring coordinated edits across multiple chart components rather than isolated fixes.
The dataset further spans varying levels of complexity, including a substantial number of challenging visually grounded editing problems (27.3\% hard problems). Difficulty labels were assigned using GPT-5 under a fixed rubric and then manually reviewed and verified by human annotators (classification prompt is in Figure~\ref{fig:difficulty_classification_prompt}).

VisEditBench spans diverse chart types, editing behaviors, and visualization paradigms. Bar and line charts are the most common chart families, reflecting their prevalence in practical visualization workflows, while the benchmark also includes scatter plots, boxplots, heatmaps, treemaps, histograms, waterfall charts, tables, and composite visualizations. 
The benchmark further covers both imperative and declarative visualization paradigms through 1,156 Matplotlib examples and 239 Vega-Lite examples. Matplotlib tasks often require procedural edits involving layout control, annotation placement, and axis formatting, whereas Vega-Lite tasks require modifying declarative specifications such as encodings, scales, legends, layers, and configuration fields. 

%% file: emnlp2020-templates/Models.tex
\subsection{Task Formulation}

We formulate visualization code editing as a multimodal code revision task. Each example is defined as \( x_i = (c_i, I_i, u_i, c_i') \), where \( c_i \) is the input visualization code, \( I_i \) is a rendered chart image, \( u_i \) is a natural-language editing instruction, and \( c_i' \) is the revised visualization code. Given \( (c_i, I_i, u_i) \), the model must generate 
executable code \( c_i' \) that produces the desired visualization while preserving the relevant data semantics.

VisEditBench supports two practical editing settings. In \textit{feedback-guided repair}, \( I_i \) contains a flawed or human-marked chart, and the model must repair the visualization according to the provided feedback. In \textit{reference-guided restyling}, \( I_i \) serves as a target chart or style reference, and the model must revise the original visualization to visually align with the reference while preserving its underlying analytical meaning.





 \begin{figure*}[t]
    \vspace{-10mm}
    \centering
    \includegraphics[width=\textwidth]{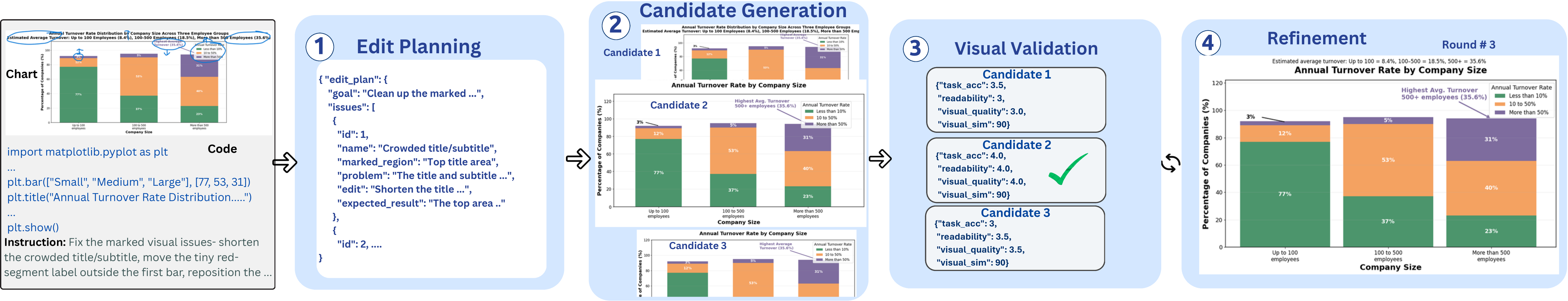}
    \caption{\textbf{VisEditAgent framework}. Given visualization code, a buggy, human-marked, or reference chart image, an instruction, and an edit intent, VisEditAgent plans the edit, generates candidate revisions, executes and renders them, validates the outputs, selects the best candidate, and refines it through iterative refinement rounds.}

    
    \label{fig:visedit_Agent}
    \vspace{-3mm}
\end{figure*} 
\vspace{-3mm}

\subsection{Model Inference} \label{subsec:data2text}

We evaluate 20 state-of-the-art VLMs spanning both closed-source and open-source model families. Closed-source models include multimodal systems from the GPT, Gemini, and Claude families \cite{openai2024gpt4technicalreport, geminiteam2024gemini15unlockingmultimodal}, while open-source models cover recent VLM and code-oriented families including Qwen, InternVL, LLaVA, and Gemma \cite{yang2024qwen2, jiang2023mistral, grattafiori2024llama, DBLP:journals/corr/abs-2406-11931, guo2025deepseek, roziere2023code}. This setup enables direct comparison between frontier proprietary systems and deployable open-source alternatives under a unified visualization editing protocol.
 We use a fixed decoding configuration for all models to support reproducibility. Full inference and evaluation parameters are provided in App.~\ref{app:inference-eval-params} (Tab. \ref{tab:inference-eval-params}).

\definecolor{closed}{RGB}{230,242,255}
\definecolor{open}{RGB}{255,239,224}

\subsection{VisEditAgent}
While zero-shot prompting can produce plausible visualization edits, real-world visualization authoring is inherently iterative~\cite{munzner2014visualization}: authors repeatedly modify code, render charts, inspect the output, and refine the visualization through successive feedback loops. This process is especially important for visually grounded edits involving layout, annotations, encodings, and style, where small code changes can trigger cascading visual errors or distort underlying data semantics. Motivated by this workflow, we introduce \textbf{VisEditAgent}, a strong render-grounded baseline 
for visualization code editing. Rather than relying on a single-pass response, VisEditAgent iteratively generates, renders, validates, and refines candidate edits using execution and visual feedback (Fig.~\ref{fig:visedit_Agent}).

\noindent\textbf{(1) Edit Planning.}
Given the input code, chart image, instruction, and editing intent, the framework first identifies the likely visual issue, relevant code regions, and required modification strategy. This stage grounds the edit in both the rendered visualization and the underlying code structure.

\noindent\textbf{(2) Candidate Generation.}
 VisEditAgent generates multiple candidate code edits representing different implementations of the requested modification. Each candidate is then executed and rendered into a chart image, enabling the framework to jointly evaluate program-level correctness and visualization-level quality. 
 Exploring multiple rendered candidates is particularly important because many visualization edits admit several plausible yet visually different solutions.

\noindent\textbf{(3) Visual Validation and Selection.}
Rendered candidates are evaluated using the task instruction, input chart, and reference or target image. The framework assesses whether each candidate satisfies the requested edit, preserves the intended data semantics, and visually aligns with the desired output. VisEditAgent then selects the strongest candidate based on execution success and visual alignment.

\noindent\textbf{(4) Feedback-Based Refinement.}
The selected candidate is further refined using execution diagnostics, rendering failures, and visual-validation feedback. This refinement stage helps correct common one-shot failure modes.

\definecolor{closedA}{RGB}{235,246,255}
\definecolor{closedB}{RGB}{220,238,252}
\definecolor{openA}{RGB}{240,248,240}
\definecolor{openB}{RGB}{225,240,225}
\definecolor{groupHeader}{RGB}{225,225,225}

\begin{table*}[h]
\centering
\scriptsize
\renewcommand{\arraystretch}{1.05}
\setlength{\tabcolsep}{3.4pt}

\caption{Pass rate by editing intent on VisEditBench. All values are percentages (\%). Overall averages all examples.} 
\vspace{-2mm}
\label{tab:viseditbench_pass_by_intent}

\resizebox{\textwidth}{!}{
\begin{tabular}{l|c c c c c c c c|c}
\toprule
\textbf{Model} &
\makecell{\textbf{Correctness}\\\textbf{Repair}} &
\makecell{\textbf{Quality}\\\textbf{Improvement}} &
\makecell{\textbf{Robustness /}\\\textbf{Generalization}} &
\makecell{\textbf{Style}\\\textbf{Adaptation}} &
\makecell{\textbf{Constraint}\\\textbf{Satisfaction}} &
\makecell{\textbf{Consistency}\\\textbf{Harmonization}} &
\makecell{\textbf{Refactor /}\\\textbf{Transformation}} &
\makecell{\textbf{Style-Aware}\\\textbf{Error Repair}} &
\makecell{\textbf{Overall}} \\
\midrule

\rowcolor{groupHeader}
\multicolumn{10}{l}{\textbf{Closed-Source Models}} \\

\rowcolor{closedA}
GPT-4o & 56.08 & 67.34 & 57.76 & 10.00 & 63.46 & 47.46 & 42.86 & 30.61 & 55.75 \\

\rowcolor{closedB}
GPT-5-Mini & 55.51 & 64.64 & 71.55 & 47.14 & 67.30 & 55.93 & 58.73 & 51.02 & 60.00 \\

\rowcolor{closedA}
GPT-5 & 61.60 & 75.45 & 56.89 & \textbf{57.14} & 59.61 & \textbf{66.10} & 69.84 & 51.02 & 65.68 \\

\rowcolor{closedB}
Gemini-3.0-Flash & 17.96 & 19.28 & 9.48 & 35.71 & 5.77 & 15.25 & 9.52 & 0.00 & 17.06 \\

\rowcolor{closedA}
Claude-4.5-Sonnet & 60.83 & 65.34 & 75.00 & 52.86 & 67.31 & 42.38 & 74.60 & 51.02 & 62.73 \\

\rowcolor{closedB}
Claude-4.6-Sonnet & \textbf{70.34} & \textbf{79.73} & \textbf{86.20} & 55.71 & \textbf{80.76} & 64.40 & \textbf{85.71} & \textbf{61.22} & \textbf{74.46} \\

\midrule

\rowcolor{groupHeader}
\multicolumn{10}{l}{\textbf{Open-Source Models}} \\

\rowcolor{openA}
Gemma-3-4B & 30.79 & 25.00 & 1.00 & 0.00 & 3.85 & 16.95 & 3.18 & 1.00 & 21.29 \\

\rowcolor{openB}
Gemma-3-12B & 42.97 & 47.97 & 6.89 & 1.43 & 13.46 & 28.81 & 7.94 & 10.20 & 34.96 \\

\rowcolor{openA}
Gemma-3-27B & 49.24 & 56.76 & 23.28 & 0.00 & 26.92 & 42.37 & 12.70 & 12.25 & 42.95 \\

\rowcolor{openB}
InternVL-3.5-1B & 11.41 & 7.88 & 0.00 & 0.00 & 1.92 & 8.47 & 1.59 & 2.04 & 7.41 \\

\rowcolor{openA}
InternVL-3.5-2B & 25.67 & 21.85 & 0.00 & 0.00 & 3.85 & 20.34 & 0.00 & 6.12 & 18.13 \\

\rowcolor{openB}
InternVL-3.5-4B & 34.03 & 33.78 & 8.62 & 0.00 & 9.61 & 28.81 & 1.58 & 2.04 & 26.33 \\

\rowcolor{openA}
InternVL-3.5-8B & 42.96 & 41.22 & 7.75 & 0.00 & 11.54 & 32.20 & 12.70 & 10.20 & 33.17 \\

\rowcolor{openB}
InternVL-3.5-14B & 41.65 & 49.78 & 3.45 & 0.00 & 3.85 & 37.29 & 7.94 & 10.21 & 34.83 \\

\rowcolor{openA}
Pixtral-12B & 44.68 & 47.07 & 5.17 & 0.00 & 15.38 & 30.51 & 6.35 & 10.20 & 35.32 \\

\rowcolor{openB}
Qwen3-VL-2B & 24.52 & 20.49 & 1.00 & 0.00 & 5.77 & 16.95 & 7.94 & 6.12 & 17.55 \\

\rowcolor{openA}
Qwen3-VL-4B & 51.11 & 50.90 & 1.72 & 1.42 & 7.69 & 37.28 & 4.76 & 16.32 & 38.84 \\

\rowcolor{openB}
Qwen3-VL-8B & 49.04 & 52.02 & 28.44 & 0.00 & 5.70 & 52.44 & 26.98 & 14.29 & 42.09 \\

\rowcolor{openA}
Qwen3-VL-30B-A3B & 49.80 & 53.38 & 44.83 & 15.71 & 38.46 & 42.37 & 47.61 & 20.41 & 47.05 \\

\rowcolor{openB}
Qwen3-VL-32B & 54.95 & 59.00 & 50.00 & 21.42 & 48.10 & 54.23 & 36.51 & 16.32 & 51.72 \\

\bottomrule
\end{tabular}
}
\vspace{-3mm}
\end{table*}


\section{Evaluation}






\subsection{Evaluation Criteria}
\vspace{-2mm}
Evaluating visualization editing requires measuring not only code executability, but also whether the edited chart satisfies the intended visual revision while remaining readable, visually coherent, and faithful to the target output. We therefore evaluate each output using five criteria: (i) \textbf{Code Execution}, which checks whether the generated code executes successfully and renders a valid chart; non-executable, timed-out, or empty outputs receive zero scores for all remaining metrics; (ii) \textbf{Task Accuracy}, which measures whether the edited chart satisfies the requested instruction or matches the target/reference chart while preserving the intended visualization semantics; (iii) \textbf{Readability and Clarity}, which evaluates the legibility and organization of labels, legends, axes, ticks, annotations, and other chart elements; (iv) \textbf{Visual Quality}, which assesses overall presentation quality, including layout, scaling, color usage, chart appropriateness, and visual polish; and (v) \textbf{Visual Similarity}, which measures how closely the rendered output aligns with the corrected or reference chart image.

\noindent\textbf{Automatic Evaluator.}
We use a rubric-based VLM evaluator with fixed prompts, deterministic settings, explicit scoring criteria, and a fixed input order. The evaluator receives 
the task instruction, the input or reference chart, and the rendered model output, without access to model identities. By default, GPT-4o serves as the evaluator for all visual metrics; when evaluating GPT-4o outputs, we instead use Gemini 2.5 Pro to avoid self-evaluation bias. We further validate the evaluator through human assessment using the same rubric and report agreement between automatic and human judgments.

\noindent\textbf{Scoring and Final Pass Rate.}
Task accuracy, readability and clarity, and visual quality are scored on a 0--5 scale with 0.5-point increments, while visual similarity is scored from 0--100. We additionally report a strict final pass rate: an example passes only if the code executes successfully, task accuracy is at least 4.5, readability and clarity and visual quality are at least 4.0, and visual similarity is at least 90. These thresholds intentionally measure complete task success rather than partial improvement, ensuring that edited charts are executable, visually coherent, instruction-faithful, and 
aligned with the intended target. The full evaluation rubric, including score definitions and intent-specific visual similarity rules, is provided in Table~\ref{tab:evaluation-rubric}.

\subsection{Main Results}
\label{sec:benchmark_results}


\subsubsection{Zero-Shot Performance} Tables~\ref{tab:viseditbench_pass_by_intent} and~\ref{tab:viseditbench_diagnostics} reveal a substantial gap between generating executable visualization code and performing visually faithful visualization editing. Overall, closed-source models substantially outperform open-source models, but even the strongest model remains far from solving visualization code editing. Claude-4.6-Sonnet achieves the best overall pass rate at 74.46\%, followed by GPT-5 at 65.68\% and Claude-4.5-Sonnet at 62.73\%. Among open-source models, Qwen3-VL-32B performs best with 51.72\%, while most open-source models remain below 50\%. Smaller models perform much worse, such as InternVL-3.5-1B at 7.41\% and Qwen3-VL-2B at 17.55\%, suggesting that visualization editing requires strong multimodal and 
reasoning ability.

\begin{table}[h!]
\centering
\small
\renewcommand{\arraystretch}{1.05}
\setlength{\tabcolsep}{5pt}

\caption{Metric-level scores on VisEditBench. Executability is reported as a percentage
, Visual Similarity on a 0--100 scale, and 
other metrics on a 5-point scale.}

\vspace{-2mm}
\label{tab:viseditbench_diagnostics}

\resizebox{\columnwidth}{!}{
\begin{tabular}{l|c c c c c}
\toprule
\textbf{Model} &
\makecell{\textbf{Code}\\\textbf{Exec.}} &
\makecell{\textbf{Task}\\\textbf{Acc.}} &
\makecell{\textbf{Read. \&}\\\textbf{Clarity}} &
\makecell{\textbf{Visual}\\\textbf{Quality}} &
\makecell{\textbf{Visual}\\\textbf{Similarity}} \\
\midrule

\rowcolor{groupHeader}
\multicolumn{6}{l}{\textbf{Closed-Source Models}} \\

\rowcolor{closedA}
GPT-4o & 93.31 & 4.04 & 4.03 & 3.95 & 78.13 \\

\rowcolor{closedB}
GPT-5-Mini & 88.20 & 4.11 & 4.12 & 4.07 & 80.65 \\

\rowcolor{closedA}
GPT-5 & 94.10 & 4.13 & 4.10 & 4.09 & 84.13 \\

\rowcolor{closedB}
Gemini-3.0-Flash & 46.02 & 1.83 & 2.07 & 2.04 & 42.85 \\

\rowcolor{closedA}
Claude-4.5-Sonnet & 94.46 & 4.41 & 4.38 & 4.38 & 86.52 \\

\rowcolor{closedB}
Claude-4.6-Sonnet & \textbf{96.19} & \textbf{4.57} & \textbf{4.54} & \textbf{4.55} & \textbf{89.42} \\

\midrule

\rowcolor{groupHeader}
\multicolumn{6}{l}{\textbf{Open-Source Models}} \\

\rowcolor{openA}
Gemma-3-4B & 82.30 & 2.70 & 3.30 & 3.33 & 64.78 \\

\rowcolor{openB}
Gemma-3-12B & 87.19 & 3.57 & 3.75 & 3.78 & 74.20 \\

\rowcolor{openA}
Gemma-3-27B & 88.27 & 3.92 & 3.97 & 3.98 & 77.34 \\

\rowcolor{openB}
InternVL-3.5-1B & 72.81 & 1.36 & 2.62 & 2.65 & 53.74 \\

\rowcolor{openA}
InternVL-3.5-2B & 80.00 & 2.39 & 3.09 & 3.13 & 61.47 \\

\rowcolor{openB}
InternVL-3.5-4B & 78.71 & 3.01 & 3.28 & 3.31 & 65.35 \\

\rowcolor{openA}
InternVL-3.5-8B & 83.38 & 3.47 & 3.64 & 3.65 & 71.25 \\

\rowcolor{openB}
InternVL-3.5-14B & 78.95 & 3.34 & 3.46 & 3.48 & 68.12 \\

\rowcolor{openA}
Pixtral-12B & 86.04 & 3.46 & 3.66 & 3.69 & 72.43 \\

\rowcolor{openB}
Qwen3-VL-2B & 79.93 & 2.24 & 3.09 & 3.13 & 62.48 \\

\rowcolor{openA}
Qwen3-VL-4B & 79.71 & 3.08 & 3.39 & 3.43 & 67.86 \\

\rowcolor{openB}
Qwen3-VL-8B & 85.83 & 3.68 & 3.79 & 3.82 & 74.77 \\

\rowcolor{openA}
Qwen3-VL-30B-A3B & 89.71 & 3.92 & 4.03 & 4.05 & 79.19 \\

\rowcolor{openB}
Qwen3-VL-32B & 87.77 & 4.05 & 4.01 & 4.04 & 78.38 \\

\bottomrule
\end{tabular}
}
\vspace{-3mm}
\end{table}
\vspace{-2mm}
A key finding is that executability alone is not the primary bottleneck for strong models. Claude-4.6-Sonnet and GPT-4o execute successfully on 96.19\% and 93.31\% of examples, respectively, yet achieve substantially lower visual similarity scores of 89.42 and 78.13. This indicates that models can often generate runnable code while still failing to preserve visual semantics, layout fidelity, or stylistic alignment with the target edit.

\begin{table*}[t]
\centering
\small
\setlength{\tabcolsep}{4pt}
\renewcommand{\arraystretch}{1.05}

\caption{\small Pass rate by editing intent for zero-shot models, VisEditAgent, and ablations. All values are percentages (\%). }
\label{tab:viseditagent_intent_results}

\resizebox{\textwidth}{!}{%
\begin{tabular}{lccccccccc}
\toprule
\textbf{Model / Method} &
\makecell{\textbf{Correct.}\\\textbf{Repair}} &
\makecell{\textbf{Quality}\\\textbf{Improve.}} &
\makecell{\textbf{Robust.}\\\textbf{Gen.}} &
\makecell{\textbf{Style}\\\textbf{Adapt.}} &
\makecell{\textbf{Constraint}\\\textbf{Satis.}} &
\makecell{\textbf{Consistency}\\\textbf{Harmon.}} &
\makecell{\textbf{Refactor/}\\\textbf{Transform.}} &
\makecell{\textbf{Style-aware}\\\textbf{Repair}} &
\textbf{Overall} \\
\midrule


\rowcolor{closedA}
GPT-4o Zero-shot & 56.08 & 67.34 & 57.76 & 10.00 & 63.46 & 47.46 & 42.86 & 30.61 & 55.75 \\

\rowcolor{closedB}
GPT-4o + VisEditAgent & 64.64 & \textbf{78.37} & \textbf{61.22} & \textbf{62.85} & \textbf{65.38} & \textbf{64.41} & \textbf{60.32} & \textbf{44.89} & \textbf{67.99} \\

\rowcolor{openA}
Qwen3-VL-4B Zero-shot & 51.11 & 50.90 & 1.72 & 1.42 & 7.69 & 37.28 & 4.76 & 16.32 & 38.84 \\

\rowcolor{openB}
Qwen3-VL-4B + VisEditAgent & 63.14 & 63.00 & 42.24 & 35.71 & 44.23 & 54.23 & 42.85 & 18.36 & 44.98 \\

\rowcolor{openA}
Qwen3-VL-8B Zero-shot & 49.04 & 52.02 & 28.44 & 0.00 & 5.70 & 52.44 & 26.98 & 14.29 & 42.09 \\

\rowcolor{openA}
Qwen3-VL-8B + VisEditAgent & \textbf{71.08} & 67.71 & 1.80 & 24.29 & 13.46 & 49.15 & 17.46 & 16.36 & 54.53 \\

\midrule

\rowcolor{groupHeader}
\multicolumn{10}{l}{\textit{Ablations on GPT-4o + VisEditAgent}} \\

\rowcolor{closedA}
w/o Multi-candidate Generation & 61.44 & 74.44 & 56.03 & 28.57 & 59.62 & 55.93 & 52.38 & 38.78 & 61.94 \\

\rowcolor{closedB}
w/o Refinement & 58.79 & 71.52 & 52.59 & 41.43 & 57.69 & 54.24 & 49.21 & 34.70 & 59.86 \\

\bottomrule
\end{tabular}%
}
\vspace{-3mm}
\end{table*}

Performance also varies sharply across editing intents. Stronger models perform relatively well on correctness repair and quality improvement, where edits often involve localized readability or layout fixes. For example, Claude-4.6-Sonnet reaches 70.34\% on correctness repair and 79.73\% on quality improvement, while GPT-4o reaches 56.08\% and 67.34\%, respectively. In contrast, visually grounded and coordinated edits remain difficult: Claude-4.6-Sonnet achieves only 55.71\% on style adaptation, and GPT-4o only 10.00\%. Overall, VisEditBench requires more than code generation or chart reconstruction: successful models must jointly reason over code, visual feedback, user intent, and rendered chart quality.

\subsubsection{VisEditAgent Results} Table~\ref{tab:viseditagent_intent_results} shows that VisEditAgent consistently improves over direct zero-shot prompting. Using GPT-4o as the base model, VisEditAgent increases overall pass rate from 55.75\% to 67.99\%, demonstrating the importance of render-grounded iterative refinement for visualization editing. The largest gains appear on visually grounded editing tasks. Style adaptation improves from 10.00\% to 62.85\%, consistency harmonization from 47.46\% to 64.41\%, and refactor/transformation from 42.86\% to 60.32\%. Similar trends hold for open-source models: for example, Qwen3-VL-4B improves from 1.42\% to 35.71\% on style adaptation and from 4.76\% to 42.85\% on refactor/transformation. Table~\ref{tab:viseditagent_diagnostics} further shows consistent improvements in task accuracy, readability, visual quality, and visual similarity. These results suggest that a central failure mode of zero-shot VLMs is not merely code generation, but the inability to iteratively validate and refine edits against rendered visual feedback. Multi-candidate generation, rendering, visual validation, and refinement substantially improve the ability of models to produce edits that are both executable and visually faithful (see Figure~\ref{fig:qa}).

To validate the automatic evaluation, we manually evaluated outputs from GPT-4o zero-shot and GPT-4o + VisEditAgent on all 1,395 VisEditBench samples, resulting in 2,790 total model outputs. We additionally evaluated Qwen3-VL-4B zero-shot and Qwen3-VL-4B + VisEditAgent on a 500-example stratified sample. Five annotators assessed each output using the same rubric as the automatic evaluator, covering task accuracy, readability and clarity, visual quality, visual similarity, and final pass/fail judgment. As shown in Table~\ref{tab:human-eval-results}, human evaluation confirms that VisEditAgent improves pass rate for both GPT-4o and Qwen3-VL-4B. The human scores also show strong agreement with the automatic evaluator, with Pearson correlations ranging from 83.38 to 87.00 across the four scalar metrics: task accuracy, readability and clarity, visual quality, and visual similarity (Table~\ref{tab:human-auto-agreement}). This suggests that the automatic evaluation is well aligned with human judgments while enabling scalable assessment of visualization code editing performance.

%% file: emnlp2020-templates/Evaluation.tex

\subsection{Ablation Studies}
\vspace{-1mm}

As shown in Tab. \ref{tab:viseditagent_intent_results}, removing multi-candidate generation lowers VisEditAgent's overall pass rate with GPT-4o from 67.99\% to 61.94\%, while removing refinement lowers it to 59.86\%. The largest drop occurs for style adaptation, where performance decreases from 62.85\% to 28.57\% and 41.43\%, respectively. These results show that both candidate selection and feedback-based refinement are important for visually faithful editing.

\vspace{-2mm}
\subsection{Error Analysis}
\label{sec:error_analysis}
\vspace{-1mm}

We conducted a qualitative error analysis on 500 randomly selected model outputs. We identify four recurring failure modes, illustrated in Fig.~\ref{fig:ea}.

\noindent\textbf{Executable but visually incorrect outputs.}
The code executes and renders a chart, but the edit is not satisfied. Typical failures include unresolved label overlap, misplaced legends, cluttered annotations, and poor readability.

\noindent\textbf{Weak grounding in visual feedback.}
Models fail to localize edits from marked or reference images. They move annotations, legends, or labels to incorrect regions, or keep arrows pointing to the wrong data points.

\noindent\textbf{Poor reference-style matching.}
In style adaptation, models often make generic visual changes rather than preserving the reference layout, colors, fonts, legends, backgrounds, or annotation style.

\noindent\textbf{Incorrect or incomplete transformations.}
For refactor/transformation tasks, models change the wrong chart type, omit required visual elements, alter encodings, or fail to preserve data semantics.

%% file: emnlp2020-templates/Appendix.tex
\appendix
\section{Appendices}
\label{app:Appendice}

\begin{table}[t!]
\centering
\small
\renewcommand{\arraystretch}{1.05}
\setlength{\tabcolsep}{5pt}

\caption{Metric-level comparison of zero-shot models and VisEditAgent. Executability is reported as a percentage (\%). Task Accuracy, Chart Readability and Clarity, and Visual Quality are reported on a 5-point scale. Visual Similarity is reported on a 0--100 scale. Best results in each column are shown in bold.}
\vspace{-2mm}
\label{tab:viseditagent_diagnostics}

\resizebox{\columnwidth}{!}{
\begin{tabular}{l|c c c c c}
\toprule
\textbf{Model / Method} &
\makecell{\textbf{Code}\\\textbf{Exec.}} &
\makecell{\textbf{Task}\\\textbf{Acc.}} &
\makecell{\textbf{Read. \&}\\\textbf{Clarity}} &
\makecell{\textbf{Visual}\\\textbf{Quality}} &
\makecell{\textbf{Visual}\\\textbf{Similarity}} \\
\midrule

\rowcolor{closedA}
GPT-4o Zero-shot & 93.31 & 4.04 & 4.03 & 3.95 & 78.13 \\

\rowcolor{closedB}
GPT-4o + VisEditAgent & \textbf{96.83} & \textbf{4.37} & \textbf{4.35} & \textbf{4.36} & \textbf{86.30} \\

\rowcolor{openA}
Qwen3-VL-4B Zero-shot & 79.71 & 3.08 & 3.39 & 3.43 & 67.86 \\

\rowcolor{openB}
Qwen3-VL-4B + VisEditAgent & 91.93 & 3.70 & 3.92 & 3.94 & 77.30 \\

\rowcolor{openA}
Qwen3-VL-8B + VisEditAgent & 93.45 & 4.03 & 4.13 & 4.16 & 76.50 \\

\bottomrule
\end{tabular}
}
\vspace{-2mm}
\end{table}

\newcommand{\pmark}{$\triangle$} 

\begin{table*}[t]
\centering
\scriptsize
\setlength{\tabcolsep}{4pt}
\begin{tabular}{p{1.9cm} p{2.1cm} p{1.8cm} c c c c c}
\toprule
\textbf{Benchmark} & \textbf{Main Task} & \textbf{Input} &
\textbf{Code Given} &
\textbf{Human Feedback} &
\textbf{Code Editing} &
\textbf{Ref. Restyling} &
\textbf{Edit Taxonomy} \\
\midrule
Text2Vis~\cite{rahman2025text2vis}
& Text-to-vis generation
& Text + data
& \pmark & \xmark & \xmark & \xmark & \xmark \\

VisEval~\cite{chen2024viseval}
& NL2VIS evaluation
& Text + data
& \xmark & \xmark & \xmark & \xmark & \xmark \\

ChartMimic~\cite{yang2025chartmimic}
& Chart-to-code generation
& Chart image
& \xmark & \xmark & \xmark & \xmark & \xmark \\

Plot2Code~\cite{wu2025plot2code}
& Plot-to-code generation
& Plot image
& \pmark & \xmark & \xmark & \xmark & \xmark \\

SWE-bench Multimodal~\cite{yang2024swe}
& Visual software repair
& Issue + repo.
& \cmark & \xmark & \xmark & \xmark & \xmark \\

\textbf{VisEditBench}
& \textbf{Visualization repair \& restyling}
& \textbf{Code + multimodal feedback}
& \cmark & \cmark & \cmark & \cmark & \cmark \\
\bottomrule
\end{tabular}
\caption{Comparison of VisEditBench with related visualization and multimodal software benchmarks. \cmark{} = yes, \xmark{} = no, and \pmark{} = partial. VisEditBench uniquely evaluates editing existing visualization code from multimodal feedback, including feedback-guided repair and reference-guided restyling.}
\label{tab:benchmark_comparison}
\end{table*}

\definecolor{gptA}{RGB}{235,246,255}
\definecolor{gptB}{RGB}{220,238,252}
\definecolor{qwenA}{RGB}{240,248,240}
\definecolor{qwenB}{RGB}{225,240,225}

\begin{table*}[t]
\centering
\scriptsize
\setlength{\tabcolsep}{4pt}
\renewcommand{\arraystretch}{1.05}

\caption{
Human evaluation results for GPT-4o and Qwen3-VL-4B under zero-shot and VisEditAgent settings. GPT-4o is evaluated on all 1,395 examples, while Qwen3-VL-4B is evaluated on a 500-example stratified sample. Code execution and pass rate are reported as percentages. Task accuracy, readability and clarity, and visual quality use a 0--5 scale; visual similarity uses a 0--100 scale.
}
\label{tab:human-eval-results}
\vspace{-2mm}

\begin{tabular*}{\textwidth}{@{\extracolsep{\fill}}lcccccc}
\toprule
\textbf{Model / Method} &
\begin{tabular}[c]{@{}c@{}}\textbf{Code}\\\textbf{Execution}\end{tabular} &
\begin{tabular}[c]{@{}c@{}}\textbf{Task}\\\textbf{Accuracy}\end{tabular} &
\begin{tabular}[c]{@{}c@{}}\textbf{Readability}\\\textbf{\& Clarity}\end{tabular} &
\begin{tabular}[c]{@{}c@{}}\textbf{Visual}\\\textbf{Quality}\end{tabular} &
\begin{tabular}[c]{@{}c@{}}\textbf{Visual}\\\textbf{Similarity}\end{tabular} &
\begin{tabular}[c]{@{}c@{}}\textbf{Pass}\\\textbf{Rate}\end{tabular} \\
\midrule

\rowcolor{gptA}
GPT-4o Zero-shot 
& 95.00 & 4.13 & 4.10 & 4.14 & 82.70 & 51.70 \\

\rowcolor{gptB}
GPT-4o + VisEditAgent 
& \textbf{97.00} & \textbf{4.36} & \textbf{4.41} & \textbf{4.40} & \textbf{89.60} & \textbf{59.10} \\

\midrule

\rowcolor{qwenA}
Qwen3-VL-4B Zero-shot 
& 80.54 & 3.11 & 3.41 & 3.44 & 68.13 & 38.91 \\

\rowcolor{qwenB}
Qwen3-VL-4B + VisEditAgent 
& \textbf{91.63} & \textbf{3.68} & \textbf{3.85} & \textbf{3.95} & \textbf{77.20} & \textbf{45.45} \\

\bottomrule
\end{tabular*}
\vspace{-3mm}
\end{table*}

\definecolor{metricBg}{RGB}{245,248,252}
\definecolor{finalBg}{RGB}{232,238,245}

\begin{table}[t]
\centering
\scriptsize
\setlength{\tabcolsep}{6pt}
\renewcommand{\arraystretch}{1.08}

\caption{
Human--automatic evaluation agreement across all human-evaluated outputs. Pearson correlations are reported for scalar metrics; agreement is reported for the final pass/fail decision.
}
\label{tab:human-auto-agreement}
\vspace{-2mm}

\begin{tabular}{p{3.8cm}c}
\toprule
\textbf{Metric} & \textbf{Pearson Correlation} \\
\midrule
\rowcolor{metricBg}
Task Accuracy & 83.38 \\
\rowcolor{metricBg}
Readability \& Clarity & 87.00 \\
\rowcolor{metricBg}
Visual Quality & 86.50 \\
\rowcolor{metricBg}
Visual Similarity & 86.00 \\
\midrule
\rowcolor{finalBg}
\textbf{Final Pass/Fail Agreement} & \textbf{81.00\%} \\
\bottomrule
\end{tabular}
\vspace{-3mm}
\end{table}

 \begin{figure*}[t]
    \vspace{-3mm}
    \centering
    \includegraphics[width=\textwidth]{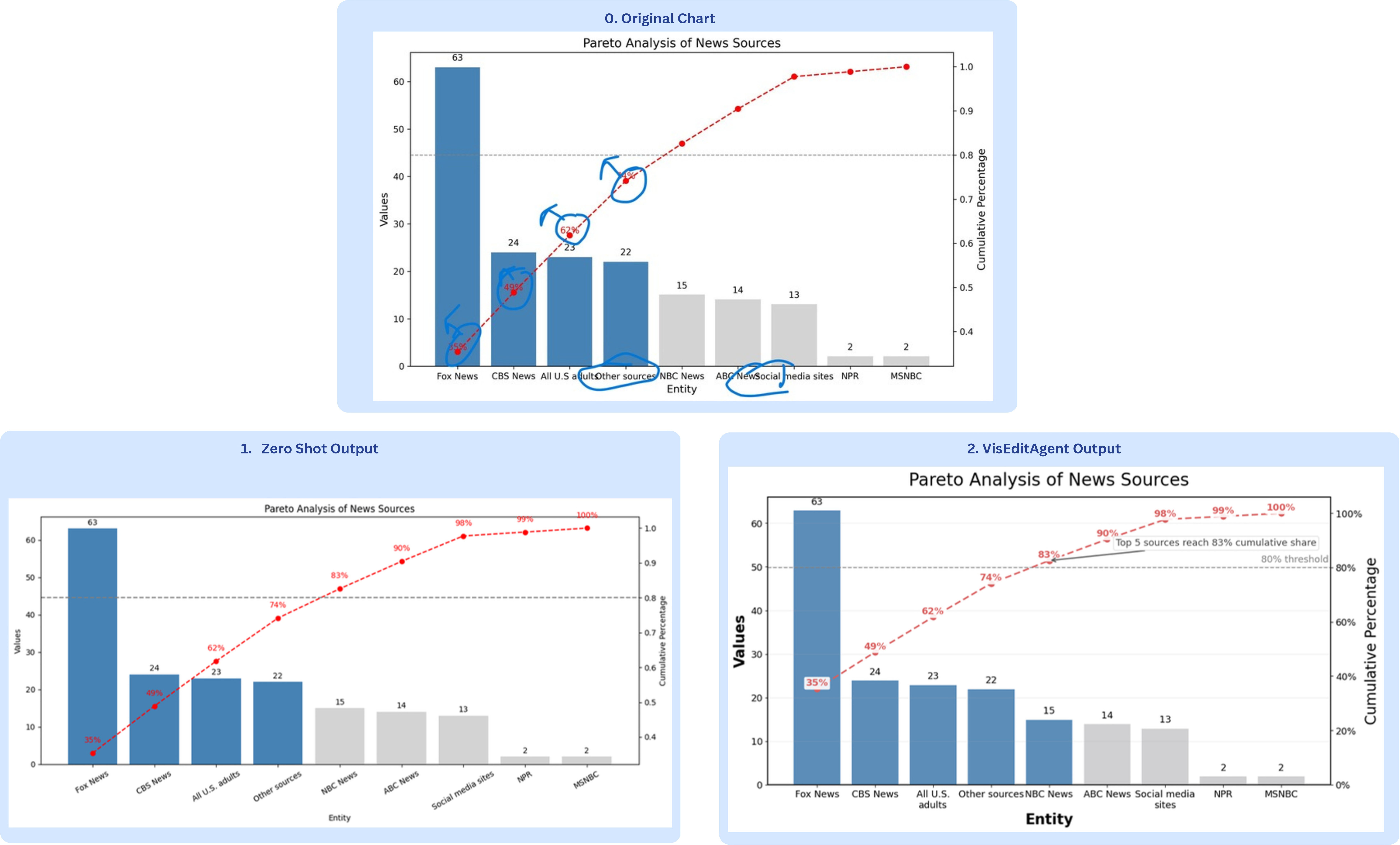}
    \caption{Qualitative example comparing GPT-4o zero-shot with GPT-4o + VisEditAgent. The original chart contains crowded x-axis labels and poorly positioned cumulative-percentage annotations. The editing instruction asks the model to improve label/annotation readability and spacing while preserving the chart semantics. Although GPT-4o zero-shot produces executable code, it only partially addresses these issues, whereas VisEditAgent produces a cleaner chart with better-positioned annotations and more readable labels.}

    \label{fig:qa}
    \vspace{-3mm}
\end{figure*} 
\vspace{-1mm}

 \begin{figure*}[t]
    \vspace{-3mm}
    \centering
    \includegraphics[width=\textwidth]{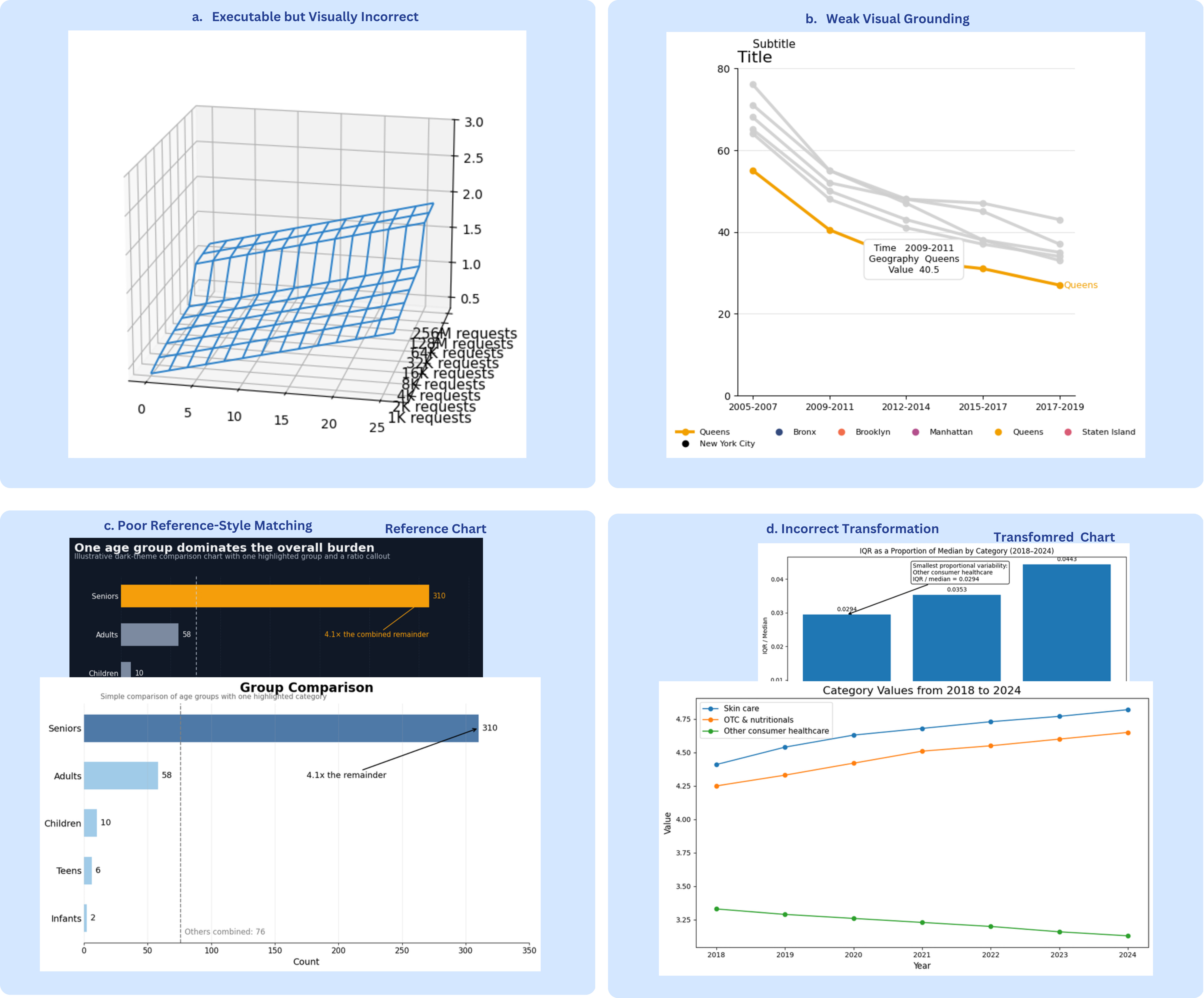}
    \caption{
Error analysis examples showing that models often produce executable and visually plausible charts while missing the intended edit. (a) GPT-4o reduces clutter but the chart remains unreadable due to overlapping 1K–256M request labels. (b) GPT-4o fails to move the legend outside the line chart, leaving it over the plotted data. (c) InternVL-14B produces a clean chart but does not match the dark reference style. (d) Qwen3-VL-4B fails to transform the output into the reference visualization.
}
    \label{fig:ea}
    \vspace{-3mm}
\end{figure*} 
\vspace{-1mm}

\begin{table*}[t]
\centering
\scriptsize
\setlength{\tabcolsep}{4pt}
\begin{tabular}{p{2.2cm} p{12.8cm}}
\toprule
\textbf{Metric} & \textbf{Rubric} \\
\midrule

\midrule

\textbf{Allowed Scores}
&
For 0--5 metrics, the evaluator may use only the following scores:
0, 0.5, 1, 1.5, 2, 2.5, 3, 3.5, 4, 4.5, and 5.
Visual similarity is scored from 0 to 100.
\\

\midrule

\textbf{Task Accuracy}
&
\textbf{Score range:} 0--5, in 0.5-point increments.

\textbf{Correctness Repair and Quality Improvement:}
5 = fully fixes the issue shown in the image and instruction, preserves data meaning, and introduces no meaningful new problems;
4.5 = almost fully fixes the issue, with only tiny remaining issues;
4.0 = mostly fixes the issue, with minor remaining problems;
0 = invalid, irrelevant, unreadable, or does not address the requested fix at all.

\textbf{All other intents:}
5 = closely matches the reference/target image and instruction, preserving data meaning unless the instruction asks otherwise;
4.5 = almost closely matches the reference/target, with only very minor differences;
4.0 = mostly matches the reference/target, with minor differences;
0 = invalid, irrelevant, unreadable, or does not match the reference/instruction at all.
\\

\midrule

\textbf{Readability \& Clarity}
&
\textbf{Score range:} 0--5, in 0.5-point increments.

5 = labels, titles, axes, ticks, legends, fonts, and annotations are clear, readable, well-positioned, and non-overlapping.

0 = no valid readable chart, or chart elements are unreadable, clipped, or overlapped so badly that the chart cannot be interpreted.
\\

\midrule

\textbf{Visual Quality}
&
\textbf{Score range:} 0--5, in 0.5-point increments.

5 = polished, professional, visually clean, accessible, uses appropriate colors/scales/chart type, and communicates the intended information clearly.

0 = no valid visualization, irrelevant image, misleading visual design, or extremely poor visual quality.
\\

\midrule

\textbf{Visual Similarity}
&
\textbf{Score range:} 0--100.

\textbf{Correctness Repair and Quality Improvement:}
Compare the rendered chart to the clean corrected version of the original buggy/marked chart. The rendered chart should remain visually similar to the original chart, but should fix the issue described in the instruction. Do not penalize the rendered chart for not reproducing handwritten marks or the original bug.

\textbf{Style Adaptation:}
Compare style similarity: colors, fonts, background, gridlines, mark style, layout, and overall appearance. Data values may differ if the task uses the base data, but the style should match the reference image closely.

\textbf{Style-Aware Error Repair:}
The reference image is the target chart style/output and may show marked potential errors. The rendered chart should recreate the clean target chart while avoiding the marked or anticipated issues. Do not reward reproducing handwritten marks or buggy artifacts.

\textbf{All other intents:}
The reference image is the target chart. The rendered chart should match it as closely as possible in chart type, data, encodings, colors, labels, title, axes, legend, annotations, layout, and overall appearance.
\\

\midrule

\textbf{Visual Similarity Scale}
&
100 = near-perfect match under the intent-specific rules;
90--99 = almost identical, only tiny differences;
80--89 = good match but minor differences;
60--79 = partially correct but noticeable differences;
40--59 = weak match;
1--39 = barely related;
0 = completely different, invalid, or no usable chart.
\\

\midrule

\textbf{Intermediate Guide for 0--5 Metrics}
&
4.5 = very good; only tiny issues.
4.0 = good; small issues but clearly successful.
3.5 = decent; mostly usable but with noticeable issues.
3.0 = average; important issues affect clarity/correctness.
2.5 = below average; several issues may mislead or confuse.
2.0 = poor; significant issues.
1.5 = very poor; major problems.
1.0 = barely usable or mostly wrong.
0.5 = almost completely failed but has a tiny relevant element.
0.0 = failed.
\\

\midrule

\textbf{Evaluator Rules}
&
If executable = 0, all scores must be 0. The evaluator is instructed to be strict. Handwritten annotations, circles, arrows, scribbles, highlights, and notes are ignored unless they describe what should be fixed. A nice-looking chart is not rewarded if it fails the instruction. A chart that changes data meaning is not rewarded unless requested.
\\

\bottomrule
\end{tabular}
\caption{
Evaluation rubric used for scoring rendered visualization outputs. Task accuracy, readability and clarity, and visual quality are scored on a 0--5 scale in 0.5-point increments. Visual similarity is scored on a 0--100 scale. The rubric is intent-aware and distinguishes repair, style adaptation, style-aware error repair, and other reference-guided target-matching tasks.
}
\label{tab:evaluation-rubric}
\end{table*}


\begin{figure*}[t]
\centering
\begin{tcolorbox}[
    colback=gray!6,
    colframe=gray!35,
    boxrule=0.5pt,
    arc=2pt,
    width=0.96\textwidth,
    left=7pt,
    right=7pt,
    top=6pt,
    bottom=6pt
]
\scriptsize
\raggedright
\setlength{\parindent}{0pt}
\setlist[itemize]{leftmargin=*, itemsep=1pt, topsep=2pt}
\setlist[enumerate]{leftmargin=*, itemsep=1pt, topsep=2pt}

\textbf{Difficulty Classification Prompt}

\medskip
You are an expert annotator for VisEditBench, a benchmark for visualization code editing from multimodal feedback.

Your task is to classify one visualization editing sample as \textbf{Easy}, \textbf{Medium}, or \textbf{Hard}.

You will be given:
\begin{enumerate}
    \item The input visualization code.
    \item The input chart image, which may show a buggy chart, a human-marked chart, or a reference/target chart.
    \item The natural-language editing instruction.
    \item Metadata such as editing intent, problem type, chart type, and visualization library, if available.
\end{enumerate}

Classify the task difficulty based on the expected effort required to produce the correct edited visualization code while preserving the intended data semantics. Use the number of distinct issues as the main signal, while also considering visual reasoning, code reasoning, issue interaction, scope of required edits, reference-style matching complexity, and risk of changing the underlying data meaning.

\medskip
\textbf{Easy.}
A task is Easy if it contains one main issue and the required edit is local, direct, and clearly specified. The issue is usually visible from the chart and can be fixed by modifying one chart property or a small number of nearby parameters. The correct code location is usually obvious, and there is little risk of changing the underlying data semantics.

Typical Easy tasks include:
\begin{itemize}
    \item increasing font size or rotating tick labels;
    \item moving a legend, title, or axis label;
    \item fixing one simple label overlap;
    \item changing a single color, marker, line style, or font property;
    \item adjusting figure size, margins, or axis-label spacing;
    \item adding or removing one simple annotation.
\end{itemize}
Choose Easy when the task mostly requires fixing one localized issue with minimal visual-code reasoning.

\medskip
\textbf{Medium.}
A task is Medium if it contains two distinct issues, or if one issue requires coordinated changes across multiple chart components. The task may require moderate reasoning over the image, instruction, and code, but the required edits are still reasonably well-scoped. Medium tasks do not need to be structurally complex; they only need to require more than one localized change or moderate visual-code alignment.

Typical Medium tasks include:
\begin{itemize}
    \item fixing two visible issues in the same chart;
    \item fixing multiple overlapping labels, ticks, or annotations;
    \item adjusting axes, legends, titles, layout, and spacing together;
    \item changing encodings while preserving the chart's meaning;
    \item improving readability across several chart components;
    \item matching a reference style with several visible properties such as colors, fonts, gridlines, and layout;
    \item modifying grouped, stacked, or multi-series charts.
\end{itemize}
Choose Medium when the task requires two fixes, several coordinated edits, or moderate visual-code reasoning, but does not require major structural transformation or highly complex semantic reasoning.

\medskip
\textbf{Hard.}
A task is Hard if it contains three or more distinct issues, multiple interacting causes, structural code changes, complex reference matching, or careful preservation of data semantics while making nontrivial visual modifications. A task can be Hard even if the instruction is short, when the correct solution requires several dependent edits or substantial reasoning about both the rendered image and the code.

Typical Hard tasks include:
\begin{itemize}
    \item fixing three or more visible issues in the same chart;
    \item converting between chart types while preserving analytical meaning;
    \item repairing a chart with multiple interacting errors;
    \item matching a complex reference style involving layout, colors, fonts, annotations, legends, background, and mark design;
    \item modifying faceted, layered, composite, or multi-panel visualizations;
    \item fixing issues caused by both data transformation and visual encoding;
    \item coordinating changes across scales, legends, annotations, encodings, layout, and data processing.
\end{itemize}
Choose Hard when the task requires three or more fixes, interacting fixes, structural transformation, complex reference matching, or high risk of changing data semantics.

\medskip
\textbf{Issue-count guideline.}
Use the number of distinct issues as the main difficulty signal, while still considering reasoning complexity. One localized issue usually suggests Easy, two distinct issues usually suggest Medium, and three or more distinct issues usually suggest Hard. However, override this rule when complexity clearly changes the difficulty: a single issue can be Medium or Hard if it requires structural code changes, complex reference matching, nontrivial data transformation, or high risk of changing data semantics; multiple issues can be lower difficulty if they are trivial, independent, and solved by the same localized edit.

\medskip
\textbf{Decision rules.}
Focus on the difficulty of producing the correct edited code, not merely understanding the instruction. Do not label a task Hard only because the chart looks visually busy, and do not label a task Easy only because the instruction is short......

\medskip
\textbf{Return only valid JSON:}
\begin{quote}
\ttfamily\scriptsize
\{\\
\hspace*{1em}"difficulty": "Easy" | "Medium" | "Hard",\\
\hspace*{1em}"explanation": "One or two sentences explaining the label."\\
\}
\end{quote}

\end{tcolorbox}
\caption{
Prompt used for difficulty classification. GPT-5 was used to classify each VisEditBench example as Easy, Medium, or Hard under this rubric.
}
\label{fig:difficulty_classification_prompt}
\end{figure*}

\subsection{Inference and Evaluation Parameters}
\label{app:inference-eval-params}

To support reproducibility, we report the decoding, rendering, and evaluation parameters used in our experiments in Table~\ref{tab:inference-eval-params}.

\begin{table*}[t]
\centering
\small
\setlength{\tabcolsep}{5pt}
\begin{tabular}{p{3.2cm} p{11.8cm}}
\toprule
\textbf{Component} & \textbf{Configuration} \\
\midrule

\textbf{Inference parameters}
&
For direct model inference, each model receives the visualization image and the corresponding editing prompt as multimodal input. The model is instructed to return only complete revised code, without markdown or explanation. Unless otherwise stated, we use image detail = \texttt{high}, temperature = 0.2, top-p = 1.0, maximum output tokens = 4096, and output format = complete executable visualization code only.
\\

\midrule

\textbf{Evaluation parameters}
&
Generated code is cleaned, executed, and rendered before scoring. Vega-Lite specifications are rendered using \texttt{vl-convert-python}, while Python/Matplotlib outputs are executed using a non-interactive \texttt{Agg} backend. Non-executable outputs are assigned executability 0 and zero scores for all remaining metrics. GPT-4o is used as the default evaluator; for GPT-4o outputs, Gemini 2.5 Pro is used to avoid self-evaluation bias. The evaluator uses temperature = 0.0, maximum output tokens = 1200, and image detail = \texttt{high}. Task accuracy, readability and clarity, and visual quality are scored on a 0--5 scale in 0.5-point increments; visual similarity is scored on a 0--100 scale.
\\

\midrule

\textbf{Final pass criteria}
&
An output is counted as a pass only if all conditions are satisfied:
\texttt{executable = 1}, task accuracy $\geq 4.5$, readability and clarity $\geq 4.0$, visual quality $\geq 4.0$, and visual similarity $\geq 90$.
\\

\bottomrule
\end{tabular}
\caption{
Inference and evaluation parameters used in VisEditBench experiments. These settings are fixed across model evaluations unless otherwise stated.
}
\label{tab:inference-eval-params}
\end{table*}